\documentclass[10pt, conference, compsocconf]{IEEEtran}
\ifCLASSINFOpdf
\else
\fi
\usepackage[cmex10]{amsmath}
\usepackage{color,graphicx}
\usepackage{bbm}
\usepackage{multirow}
\usepackage{rotating}
\usepackage[table]{xcolor}
\usepackage{hhline,longtable,booktabs}
\usepackage{pifont}
\newcommand{\cmark}{\ding{51}}%
\newcommand{\xmark}{\ding{55}}%

\begin{document}
%
\title{Reducing the Effects of Detrimental Instances}

\author{\IEEEauthorblockN{Michael R. Smith}
\IEEEauthorblockA{Department of Computer Science\\
Brigham Young University\\
Provo, UT 84602\\
Email: msmith@axon.cs.byu.edu}
\and
\IEEEauthorblockN{Tony Martinez}
\IEEEauthorblockA{Department of Computer Science\\
Brigham Young University\\
Provo, UT 84602\\
Email: martinez@cs.byu.edu}}


%


\maketitle

\begin{abstract}
Not all instances in a data set are equally beneficial for inducing a model of the data.
Some instances (such as outliers or noise) can be detrimental.
However, at least initially, the instances in a data set are generally considered equally in machine learning algorithms.
Many current approaches for handling noisy and detrimental instances make a binary decision about whether an instance is detrimental or not.
In this paper, we 1) extend this paradigm by weighting the instances on a continuous scale and 2) present a methodology for measuring how detrimental an instance may be for inducing a model of the data.
We call our method of identifying and weighting detrimental instances \textit{reduced detrimental instance learning} (\textit{RDIL}).
We examine RDIL on a set of 54 data sets and 5 learning algorithms and compare RDIL with other weighting and filtering approaches.
RDIL is especially useful for learning algorithms where every instance can affect the classification boundary and the training instances are considered individually, such as multilayer perceptrons trained with backpropagation (MLPs).
Our results also suggest that a more accurate estimate of which instances are detrimental can have a significant positive impact for handling them.
\end{abstract}


%
\IEEEpeerreviewmaketitle

\section{Introduction}
The goal of supervised machine learning is to induce an accurate generalizing function from a set of labeled training instances.
Given that in most cases, all that is known about a task is contained in the set of training instances, at least initially, the instances in a data set are generally considered equally.
However, some instances are more beneficial than others for inducing a model of the data. 
For example, outliers or mislabeled instances are not as beneficial as border instances and can even be detrimental in many cases.
In addition, other instances can be detrimental for inducing a model of the data even if they are labeled correctly and are not outliers.



\begin{figure}
\centering
\begin{tabular}{|c|c|}
\hline
\input{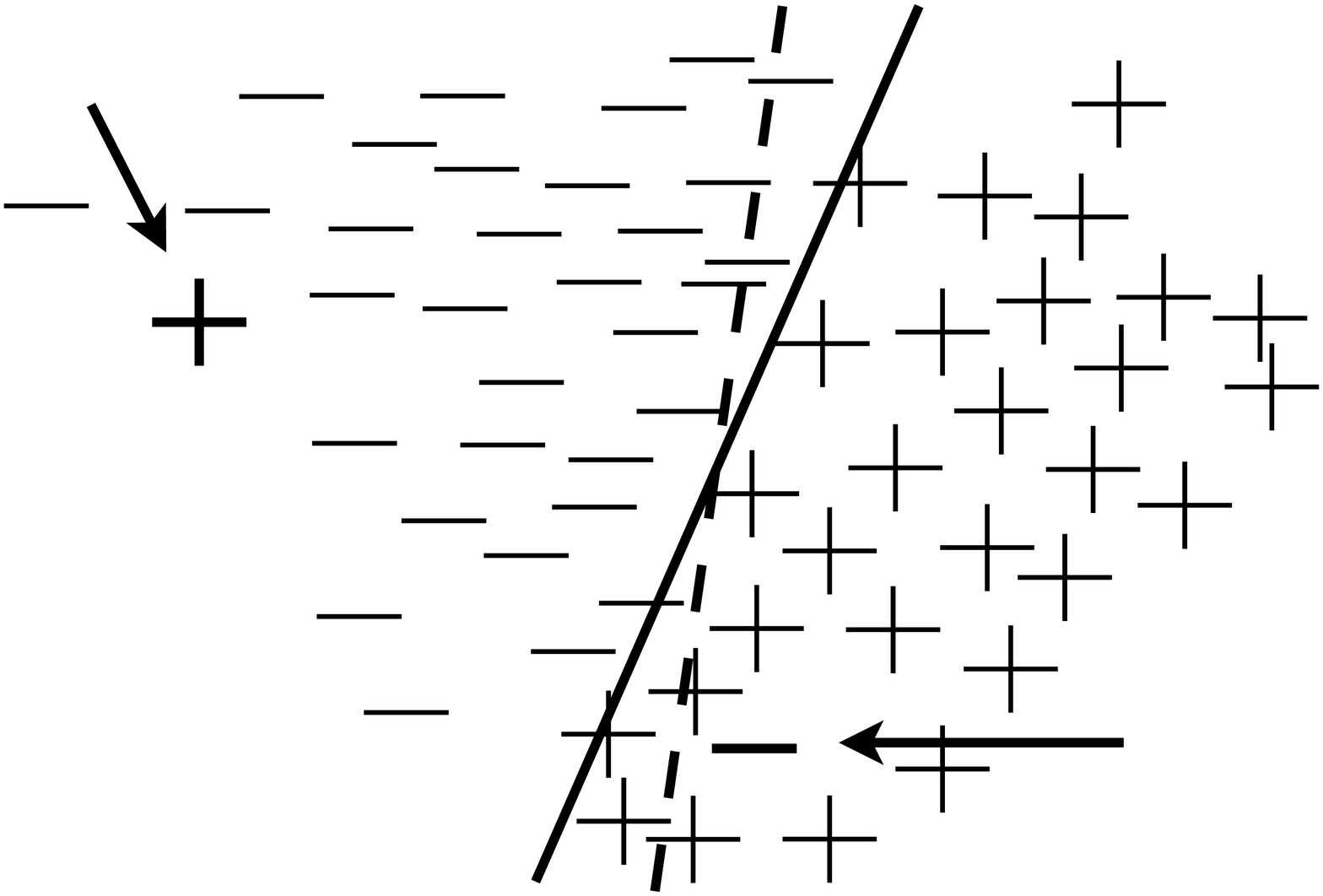}  & 
\input{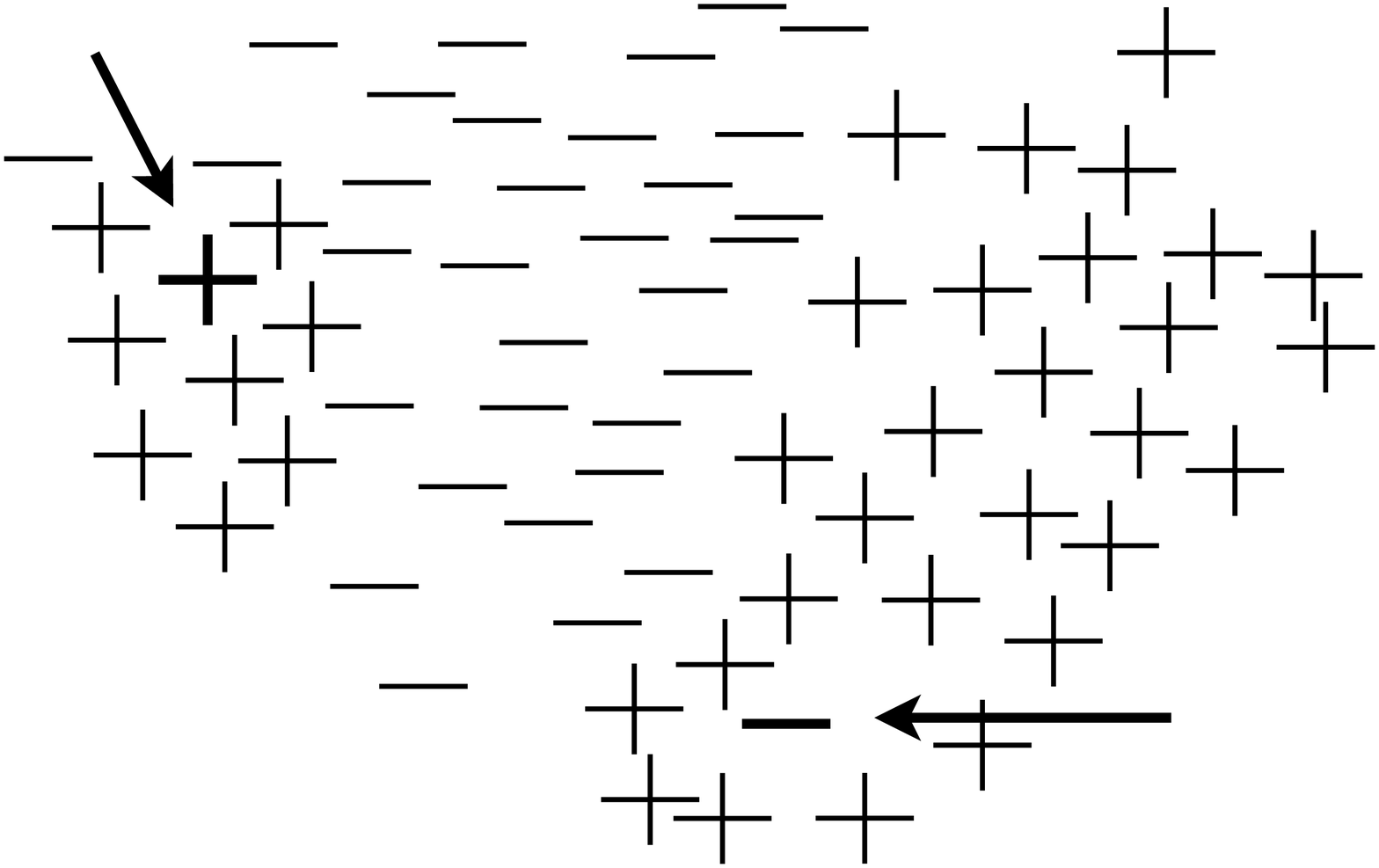}\\
\hline
\multicolumn{1}{c}{a}&\multicolumn{1}{c}{b}\\
\end{tabular}
\caption{Hypothetical, 2-dimensional data set with two detrimental instances (instances A and B) that shows a) that treating all instances equally in a data set with detrimental instances can adversely affect the classification boundary and b) that how detrimental an instance may be is dependent on the other instances in a data set.}
\label{figure:example}
\end{figure}

A possible effect of considering all instances equally, including the detrimental instances, when inducing a model of the data is shown in the hypothetical two-dimensional data set in Figure \ref{figure:example}a.
The solid line represents the ``actual'' classification boundary and the dashed line represents a potential induced classification boundary.
Instances A and B are detrimental instances that ``pull'' the decision boundary away from the true boundary and cause the instances in the space between the true boundary and induced boundary to be misclassified.
A learning algorithm can more precisely model the data by considering instances differently during training to suppress the effects of detrimental training instances.

This is especially true for learning algorithms such as backpropagation for training multilayer perceptrons (MLP).
Detrimental instances (e.g. instance A) have the greatest effect on the classification boundary since they can have the largest error value.
As shown by Elman \cite{Elman1993}, this is particularly important during the early stages of training a MLP when the initial gradient is calculated.
Elman proposes a method to initially train the MLP with simpler instances then gradually increase to more complex instances.
This procedure has developed into curriculum learning and has had success in deep learning \cite{Bengio2009}.
However, it is not as successful for shallow MLPs \cite{Smith_CL}.
The impact of detrimental instances is lessened in other other learning algorithms since the influence of a particular instance is localized.
For example, \textit{k}-NN only considers the \textit{k} nearest neighbors of an instance.

Assuming that all that is known about a task is contained in the training set, how detrimental an instance is for inducing a model of the data can be estimated based on its relationship with the other instances in the data set.
For example, instance A from Figure \ref{figure:example}a represents a detrimental instance as an outlier--being in a region with instances of a differing class.
In contrast, instance A in the data set shown in Figure \ref{figure:example}b is not as detrimental given additional instances of the same class in the same region.
%
As determining if an instance is detrimental exhibits a degree of uncertainty, we examine weighting the instances in a data set by their likelihood of being misclassified.
Instance A from Figure \ref{figure:example}a, for example, has a high likelihood of being misclassified while instance B may have a lower likelihood of being misclassified.
Weighting the instances limits the influence of an instance proportionate to its detrimentality measure.
We present a theoretically-motivated methodology for estimating the likelihood that an instance will be misclassified that lessens the dependence on any one model.
We call this approach of weighting the instances by their probability of being misclassified \textit{reduced detrimental instance learning} (\textit{RDIL}).
Filtering or removing detrimental instances prior to training can be viewed as a special case of instance weighting.
We show that both filtering and weighting are viable solutions and examine when each is most beneficial.

We examine RDIL on a set of 5 learning algorithms and 54 data sets.
We compare multiple versions of RDIL with another weighting scheme, pair-wise expectation maximization (PWEM) \cite{Rebbapragada2007}, as well as several filtering algorithms: misclassification filters and an ensemble filter \cite{Brodley1999}, and repeated-edited nearest-neighbor (RENN) \cite{Tomek1976}.
We find that some learning algorithms benefit more from filtering and others from instance weighting.
Specifically, filtering is more beneficial for decision trees, rule-based learners, and nearest neighbor algorithms.
Weighting the instances has a more significant impact on multilayer perceptrons.
In cases with high amounts of noise, weighting the instances is generally preferable to filtering as it generally achieves higher accuracy and does not require a threshold to be set if filtering is based on a continuous value.

The remainder of the paper is organized as follows.
Section \ref{section:relatedWorks} reviews related work in handling noise. 
Section \ref{section:math} motivates weighting the instances.
The detrimentality measure is presented in Section \ref{section:estimatingP}.
Our experimental methodology is presented in Section \ref{section:methodology}.
The results of RDIL are provided in Section \ref{section:results}.
Section \ref{section:conclusions} concludes the paper.

\section{Related Work}
\label{section:relatedWorks}
Many real-wold data sets contain detrimental instances that arise from noise (typos, measurement errors, etc.) or from the stochastic nature of the task.
Noise is a subset of detrimental instances.
Previous work has examined how class noise and attribute noise affects the performance of various learning algorithms~\cite{Nettleton2010}. 
They found that class noise is generally more harmful than attribute noise.
The consequences of class noise, as summarized by Fr\'{e}nay and Verleysen \cite{Frenay2014}, include 1) a deterioration of classification performance, 2) increased learning requirements and model complexity, and 3) a distortion of observed frequencies.

Most learning algorithms are designed to tolerate a certain degree of detrimental instances by making a trade-off between the complexity of the induced model and minimizing error on the training data to prevent overfit.
For example, to avoid overfit many algorithms use a validation set for early stopping and/or regularization by adding a complexity penalty to the loss function \cite{bishop2006pattern}.
Some learning algorithms have been adapted specifically to better handle noise.
Boosting algorithms \cite{Schapire1990}, for example, assign more weight to misclassified instances--which often include mislabeled and noisy instances.
To address this, Servedio \cite{Servedio2003} presented a boosting algorithm that does not place too much weight on any training instance. 

Preprocessing the data set explicitly handles detrimental instances by removing, weighting, or correcting them.
Filtering detrimental instances has received much attention and has generally been shown to result in an increase in classification accuracy, especially when there are large amounts of noise \cite{Gamberger2000,Smith2011}.
One frequently used filtering technique removes any instance that is misclassified by a learning algorithm \cite{John95} or set of learning algorithms \cite{Brodley1999}.
Other approaches use information theoretic or machine learning heuristics to remove noisy instances.
Segata et al. \cite{Segata2009}, for example, remove instances that are too close or on the wrong side of the decision surface generated by a support vector machine.
However, filtering has the potential downside of discarding useful instances and/or too many instances.

Rather than making a binary decision about the detrimentality of in instance, weighting allows a continuous scale.
Filtering can be considered a special case of weighting where each instance is assigned a weight of 0 or 1.
Rebbapragada and Brodley \cite{Rebbapragada2007} weight the instances using expectation maximization to cluster instances that belong to a pair of the classes.
The probabilities between classes for each instance is compiled and used to weight the influence of each instance.

Data cleaning does not discard any instances, but rather strives to correct the noise in the instances.
As in filtering, the output from a learning algorithm has been used to clean the data.
Polishing \cite{Teng2003} trains a learning algorithm (in this case a decision tree) to predict the value for each attribute (including the class).
The predicted (i.e.~corrected) attribute values for the instances that increase generalization accuracy on a validation set are used instead of the uncleaned attribute values.


\section{Modeling Detrimentality}
\label{section:math}

\begin{figure*}
\begin{center}
\begin{tabular}{|c|c||c|c|}
\hline
\includegraphics[scale=0.85]{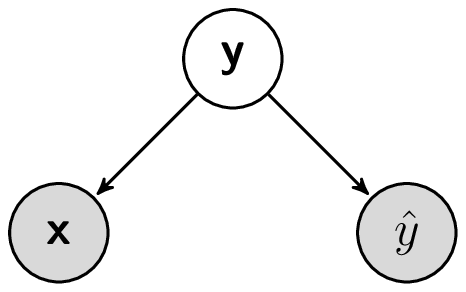} & \includegraphics[scale=0.85]{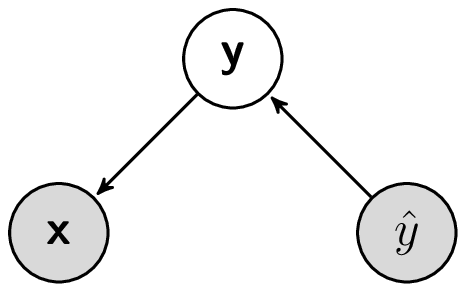} & \includegraphics[scale=0.85]{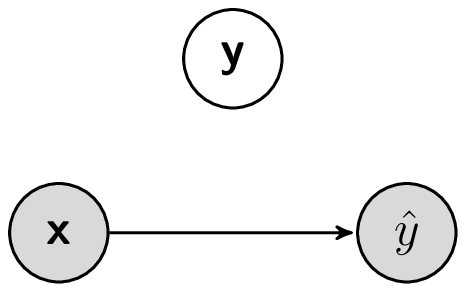} & \includegraphics[scale=0.85]{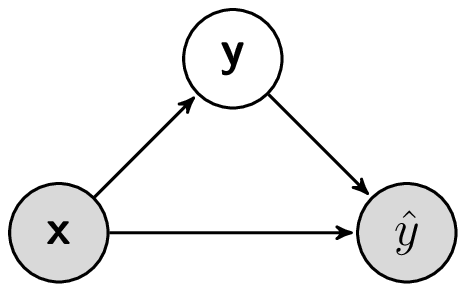} \\
\hline
\multicolumn{1}{c}{a} & \multicolumn{1}{c}{b}& \multicolumn{1}{c}{c}& \multicolumn{1}{c}{d}\\
\end{tabular}
\caption{Graphical model of a a) generative probabilistic model, b) the generative model proposed by Lawrence and Sch\"{o}lkopf \cite{Lawrence2001} and a discriminative probabilistic model for c) $p(\hat{y}|x)p(x)$ and d) $p(\hat{y}|x,y)p(y|x)p(x)$.}
\label{figure:generativeModel}
\end{center}
\end{figure*}

Lawrence and Sch\"{o}lkopf \cite{Lawrence2001} model a data set using a generative model that also models the noise.
Let $T$ be a training set composed of instances $\langle x_i, \hat{y}_i\rangle$ drawn i.i.d. from the underlying data distribution $\mathcal{D}$.
Each instance has an associated latent random variable/feature $y_i$.
Thus, $x_i$ is the set of input features, $\hat{y}_i$ is the possibly noisy class label given in the training set, and $y_i$ is the true unknown class label.
Lawrence and Sch\"{o}lkopf assume that the joint distribution $p(x_i,y_i,\hat{y}_i)$ is factorized as $p(\hat{y}_i|y_i)p(x_i|y_i)p(y_i)$ as shown in Figure \ref{figure:generativeModel}a.
Since modeling the prior distribution of the unobserved random variable $y_i$ is not feasible, they estimate the prior distribution of $p(\hat{y}_i)$ with some assumptions about the noise as shown in Figure \ref{figure:generativeModel}b.

Following the premise of Lawrence and Sch\"{o}lkopf, we explicitly model the possibility that an instance is mislabeled (i.e. $y \neq \hat{y}$).
Rather than using a generative model, though, we use a discriminative model since we are focusing on classification tasks and do not require the full joint distribution.
Also, discriminative models have been shown to yield better performance on classification tasks \cite{Ng+Jordan:2001}.
A generative model, which models the full joint distribution of $p(x, y, \hat{y})$, differs from a discriminative model which is mostly concerned with modeling the likelihood of the class: $p(\hat{y}|x)$.
%
%
Given a training set $T$, a discriminative model generally seeks to find the most probable hypothesis $h$ that maps each $x_i \mapsto \hat{y}_i$--ignoring the fact that $\hat{y}_i$ may not equal $y_i$.
This is shown graphically in Figure \ref{figure:generativeModel}c where $p(\hat{y}_i | x_i) p(x_i)$ is estimated using a discriminative approach such as a neural network or a decision tree to induce a hypothesis of the data.
The possibility of label noise is not explicitly modeled in this form (i.e. $p(y_i)$ is ignored).
Label noise is generally handled by avoiding overfit such that more probable, simpler hypotheses are preferred.

Label noise can be more explicitly modeled by considering that $\hat{y}_i$ may be noisy by associating a latent random variable $y_i$ with each instance. 
In this context, a supervised learning algorithm seeks to maximize $p(\hat{y}_i|x_i,y_i)p(y_i|x_i)p(x_i)$--modeled graphically in Figure \ref{figure:generativeModel}d.
This factorization of the likelihood for the observed class label for an instance suggests that it should be weighted by the probability that $y_i$ is the actual class.
Thus, when considering the possibility that $\hat{y}_i \neq y_i$, it is natural to weight the instances by $p(y_i|x_i)$.
This provides the motivation for \textit{reduced detrimental instance learning} (RDIL).
RDIL takes two passes through the data set.
In the first pass, $p(y_i|x_i)$ is calculated.
Next, a learning algorithm is trained with the training instances weighted by $p(y_i|x_i)$.
However, calculating $p(y_i|x_i)$ is not trivial.
A method to estimate $p(y_i|x_i)$ is described in the following section.

\section{Estimating $p(y_i|x_i)$}
\label{section:estimatingP}

In this paper, $p(y_i|x_i)$ is used as the detrimentality measure for each training instance.
In general, $p(y_i|x_i)$ does not make sense outside the context of an induced hypothesis.
Thus, using an induced hypothesis $h$ from a learning algorithm trained on $T$, the quantity $p(y_i|x_i,h)$ can be approximated as $p(\hat{y}_i|x_i,h)$ assuming that $p(y_i|\hat{y}_i)$ is represented in $h$.
In other words, the induced discriminative model is able to model if one class label is more likely than another class label given an observed noisy label.
After training a learning algorithm on $T$, the class distribution for an instance $x_i$ can be calculated based on the output from the learning algorithm.

The dependence of $p(y_i|x_i)$ on a particular hypothesis $h$ can be removed by summing over all possible hypotheses $h$ in $\mathcal{H}$ and multiplying each $p(\hat{y}_i|x_i,h)$ by $p(h)$:
\begin{equation}
p(y_i|x_i)\approx p(\hat{y}_i|x_i) = \sum_{h\in\mathcal{H}} p(\hat{y}_i|x_i,h)p(h). \label{eq:sumOverH}
\end{equation}
This formulation is infeasible, though, because 1) it is not practical (or possible) to sum over the set of all hypotheses, 2) calculating $p(h)$ is non-trivial, and 3) not all learning algorithms produce a probability distribution.
These limitations make probabilistic generative models attractive, such as the kernel Fisher discriminant algorithm \cite{Lawrence2001}.
However, for classification tasks, discriminative models generally have a lower asymptotic error than generative models \cite{Ng+Jordan:2001}.

\begin{figure}
\begin{center}
\vskip -2mm
\input{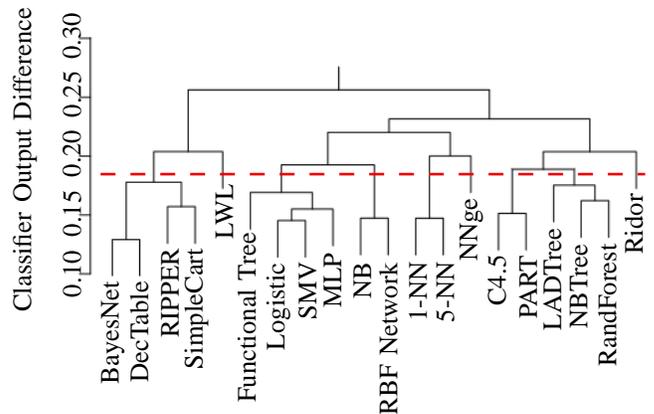}

\caption{Dendrogram of the considered learning algorithms clustered using unsupervised metalearning based on their classifier output difference.}
\label{figure:COD}
\end{center}
\end{figure}

In this paper we approximate $p(\hat{y}_i|x_i)$ using a diverse subset of $\mathcal{H}$.
The diversity of the subset of $\mathcal{H}$ refers to a set of hypotheses that differ in their classification of novel instances.
The diverse subset of $\mathcal{H}$ is created using unsupervised meta-learning (UML) \cite{Lee2011}.
UML first uses classifier output difference (COD) \cite{Peterson2005} to measure the diversity between learning algorithms.
COD measures the distance between two learning algorithms as the probability that the learning algorithms make different predictions.
UML then clusters the learning algorithms based on their COD scores with hierarchical agglomerative clustering.
We considered 20 learning algorithms from Weka with their default parameters \cite{weka2009}.
The resulting dendrogram is shown in Figure \ref{figure:COD}, where the height of the line connecting two clusters corresponds to the distance (COD value) between them.
A cut-point of 0.18 was chosen to create 9 clusters and a representative algorithm from each cluster was chosen to create a diverse set of $\mathcal{H}$.
Other numbers of clusters could have been used.
The learning algorithms that are used to estimate $p(\hat{y}_i|x_i)$ are listed in Table \ref{table:LA}.

\begin{table}
\caption{The diverse set of algorithms (as determined by unsupervised meta-learning) used to estimate $p(y_i|x_i)$.}
\label{table:LA}
\centering
\begin{tabular}{ll}
\hline
\multicolumn{2}{c}{Learning Algorithms}\\
\hline
*& Multilayer Perceptron trained with Back Propagation (MLP) \\
*& Decision Tree (C4.5) \\
*& Locally Weighted Learning (LWL) \\
*& 5-Nearest Neighbors (5-NN) \\
*& Nearest Neighbor with generalization (NNge) \\
*& Na\"{i}ve Bayes (NB) \\
*& RIpple DOwn Rule learner (RIDOR) \\
*& Random Forest (Random Forest) \\
*& Repeated Incremental Pruning to Produce Error Reduction (RIPPER) \\
\end{tabular}
\end{table}

$p(\hat{y}_i|x_i)$ is estimated for each training instance using 10-fold cross-validation (the instance $\langle x_, \hat{y}_i \rangle$ is \textit{not} used to induce the hypothesis $h$).
Using a set of diverse hypotheses induced by the learning algorithms in $\mathcal{L}$, we approximate $p(\hat{y}_i|x_i)$ as:
\begin{align}
p(\hat{y}_i|x_i) \approx p(\hat{y}_i|x_i,\mathcal{L}) 
&= \frac{1}{|\mathcal{L}|} \sum_{j=1}^{|\mathcal{L}|} p(\hat{y}_i|x_i, l_j(T)) \label{eq:final}
\end{align}
where $l_j(T)$ is the hypothesis induced by the $j^{th}$ learning algorithm trained on $T$.
From Equation \ref{eq:sumOverH}, $p(h)$ is estimated as $\frac{1}{|\mathcal{L}|}$ for the hypotheses induced by the learning algorithms in $\mathcal{L}$ and as zero for all of the other hypotheses in $\mathcal{H}$.

\section{Methodology}
\label{section:methodology}
We investigate the effects of filtering and weighting on the C4.5, 5NN, MLP, Random Forest, and RIPPER learning algorithms (abbreviated in Table \ref{table:LA}). 
Table \ref{table:QW} summarizes how an instance is weighted by $p(y_i|x_i)$ for the examined learning algorithms.
For MLPs trained with backpropagation, the error ($(t-o)f'(net)$) is scaled by $p(y_i|x_i)$ where $(t-o)$ is the difference between the target value and the output of the network, $f'(net)$ is the derivative of the activation function $f$ and $net$ is the sum of the product of each input $i_j$ and its corresponding weight $w_j$: $net=\sum_j w_ji_j$.
For Random Forests, the distribution for selecting instances in the random trees is weighted by $p(y_i|x_i)$ rather than being uniformly weighted. 
For the other learning algorithms that keep track of counts, each instance is weighted by $p(y_i|x_i)$.
$\sum_{c_i}$ represents summing over instances that meet some criterion $c_i$ and $\sum_T$ sums over all of the instances in the data set.

\begin{table}[t]
\centering
\caption{How instance weighting is integrated into the considered learning algorithms.}
\setlength{\tabcolsep}{2pt}
\begin{tabular}{l|c|c}
LA & Orig & RDIL\\
\hline
MLP & $(t-o)f'(net)$ & $p(y_i|x_i)(t-o)f'(net)$\\
\hline
Random Forest & Uniform dist & Weighted by $p(y_i|x_i)$\\
\hline
C4.5,& Count number of instances, i.e.  & Sum $p(y_i|x_i)$ \\
5-NN, & \multirow{2}{*}{$\frac{\sum_{c_i} 1}{\sum_T 1}$} & \multirow{2}{*}{$\frac{\sum_{c_i} p(y_i|x_i)}{\sum_T p(y_i|x_i)}$}\\
RIPPER&  & \\
& &\\
\end{tabular}
\label{table:QW}
\end{table}

We consider three weighting schemes: RDIL-$\mathcal{L}$, RDIL-Biased, and PWEM described below.
1) \textbf{RDIL-$\mathcal{L}$} uses $p(\hat{y}_i|x_i, \mathcal{L})$. 
Since not all learning algorithms in Table \ref{table:LA} produce a probability distribution, the Kronecker delta function $\delta(g(x_i),y_i)$ is used in this paper instead of $p(\hat{y}_i|x_i,h)$ where $h(x_i)$ returns the predicted class from the induced hypothesis $h$ given input features $x_i$.
2) \textbf{RDIL-Biased} approximates $p(y_i|x_i)$ as $p(\hat{y}_i|x_i,h)$ where the hypothesis $h$ is induced by the same learning algorithm that is used to induce a model of the data.
To get a real-value from a single hypothesis, we compute a classifier score for each instance from the learning algorithm.
Below, we present how we calculate the classifier scores for the investigated learning algorithms.

\noindent {\bf MLP:}
For multiple classes, each class from a data set is represented with an output node.
After training a MLP with backpropagation, the classifier score is the largest value of the output nodes normalized between zero and one: 
$
\hat{p}(y | x) = \frac{o_i(x)}{\sum_i^{|Y|} o_i(x)}
$
where $y$ is a class from the set of possible classes $Y$ and $o_i$ is the value from the output node corresponding to class $y_i$.

\noindent {\bf C4.5:}
To calculate a classifier score, an instance first follows the induced set of rules until it reaches a leaf node.
The classifier score is the number of training instances that have the same class as the examined instance divided by all of the training instances that also reach the same leaf node.

\noindent {\bf 5-NN:}
The percentage of the nearest-neighbors that agree with the class label of an instance as the classifier score.

\noindent {\bf Random Forest:}
For each tree, an instance follows the induced set of rules until it reaches a leaf node.
The counts from the reached leaf nodes for each class are summed together and then normalized between 0 and 1.

\noindent {\bf RIPPER:}
The percentage of training instances that are covered by a rule and share the same class as the examined instance.


\noindent Obviously, a classifier score does not produce a true probability.
However, the classifier scores approximate the confidence of $p(y_i|x_i)$.
3) Pair-wise expectation maximization (\textbf{PWEM}) \cite{Rebbapragada2007} weights each instance using the EM algorithm.
For each pair of classes, the instances that belong to the two classes are clustered using EM where the number of clusters is determined using the Bayesian Information Criterion \cite{Kass1995}. 
Given the $Y-1$ clusterings ($Y$ is the number of classes in the data set), $p(y|x)$ is calculated as:
\thinmuskip=0mu
\begin{align}
 p(y_i|x_i)\! &= \!\sum_\theta p(\theta)p(y_i|x_i, \theta)\!= \!\sum_\theta p(\theta)\!\sum_{c=1}^k p(y_i|c,\theta)p(c|x_i,\theta) \nonumber
\end{align}
where $\theta$ is a clustering model induced using the EM algorithm, $c$ is a cluster in $\theta$, and $k$ is the number of clusters in $\theta$.

We also compare weighting with three filtering techniques.
1) \textbf{Filter-$\mathcal{L}$} uses $p(\hat{y}_i|x, \mathcal{L})$ for filtering, similar to the three learning algorithm ensemble filter examined by Brodley and Friedl \cite{Brodley1999}.
Instances that are misclassified by 50\% of the learning algorithms in the ensemble are filtered from the training set. 
Note that other percentages could also be used.
We found that 50\% generally produces good results compared to values of 70\% and 90\%.
In practice a validation set could be used to determine the percentage that would be used.
2) {\bf Filter-Biased} removes any instance that is misclassified by the same learning algorithm that is being used to induce a model from the training set.
3) Repeated-edited nearest-neighbor (\textbf{RENN}) \cite{Tomek1976} repeatedly removes the instances that are misclassified by a 3-nearest neighbor classifier.

Each noise handling method is evaluated by averaging the results from ten runs of each experiment. 
For each experiment, the data is shuffled and then split into 2/3 for training and 1/3 for testing.
The training and testing sets are stratified.
Random noise is introduced by randomly changing $n\%$ of the training instances to a new label chosen uniformly from the possible class labels (noisy completely at random).
The noise levels are examined at 0\%, 10\%, 20\%, 30\%, and 40\%.
We examine noise handling using the 5 chosen learning algorithms on a set of 54 data sets from the UCI data repository \cite{UCI_dataRepo}.
Statistical significance between pairs of algorithms is determined using the Wilcoxon signed-ranks test as suggested by Dem\v{s}ar \cite{Demsar2006}.

As there is no way to determine if an instance is noisy or mislabeled without the use of a domain expert, most previous work adds artificial noise to show the impact of noise and how handling noise improves the accuracy.
Generally, once there are large amounts of noise, a noise handling approach significantly increases the classification accuracy.
In the following experiments, artificial noise is added to the data sets.

\section{Results}
\label{section:results}
In this section, we present the results of our experiments.
For the tables in this section, the algorithm in the first row is the baseline algorithm that the algorithms in the subsequent rows are compared against.
The values in the ``g,e,l'' rows represent the number of times that the accuracy from the baseline algorithm is greater than, equal to, or less than the compared algorithm.
A \xmark~represents cases where the baseline algorithm achieves significantly higher classification accuracy.
A \cmark~represents cases where the accuracy from the compared algorithm is significantly higher than the baseline algorithm.

Table \ref{table:noNoise} compares no noise handling (Orig) with the considered noise handling techniques. 
The only noise handling technique that significantly increases classification accuracy is a MLP using RDIL-$\mathcal{L}$.
In contrast, no noise handling achieves significantly higher accuracy than using a noise handling technique in several cases.
RENN achieves significantly lower classification accuracy for all of the considered learning algorithms.
This highlights a point that is often overlooked in the noise handling literature--noise handling can be detrimental if used in all cases.
Previous work has generally considered only a few data sets where noise handling is beneficial.
The impact of filtering or weighting is also dependent on which learning algorithm is used to induce a model of the data.
As expected, MLPs achieve the most significant increase in accuracy with instance weighting.
On the other hand, C4.5, 5-NN, Random Forests, and RIPPER achieve the most significant increase in accuracy with filtering.

\begin{table}
\centering
\caption{The average accuracy over the 54 data sets for the 5 considered learning algorithms using the investigated noise handling approaches with no artificial noise added to the data sets.
A \cmark~to the left represent cases where the noise handling approach significantly increases the accuracy and \xmark~where the noise handling approach significantly decreases the accuracy.}
\setlength{\tabcolsep}{3.0pt}
\begin{tabular}{l|ccccccccccc}
& C4.5& & 5-NN & & MLP & & Rand For & & RIPPER \\ 
\hline
Orig & 79.31 & & 79.37 && 81.67 && 81.18 && 78.35 \\ 
\hline
RDIL-$\mathcal{L}$ & 78.19 & & 78.72 & & \textbf{82.26} & \cmark & 80.82 & & 77.86 & \\ 
g,e,l & 27,1,26& & 27,4,23 & & 18,3,33 & & 28,2,24 & & 26,2,26 \\ 
\hline
RDIL-Biased & 79.29 &  & 78.34 & \xmark & 81.49 &  & 80.94 &  & 77.98 &  \\ 
g,e,l & 23,7,24 & & 32,7,15 & & 23,5,26 & & 26,4,24 & & 29,4,21 \\ 
\hline
PWEM & 76.41 & & 78.02 & \xmark & 82.79 & & 81.51 & & 74.17 & \\ 
g,e,l & 30,3,21 & & 33,3,18 & & 23,3,28 & & 34,1,19 & & 39,1,14 \\ 
\hline
Filter-$\mathcal{L}$  & 79.55 & & 79.40 & & 81.80 & & 81.66 & & 78.98 & \\ 
g,e,l & 25,11,18 & & 23,9,22 & & 23,4,27 & & 28,2,24 & & 27,5,22\\ 

\hline
Filter-Biased & 79.34 & & 76.99 & \xmark & 81.39 & & 81.16 & & 77.20 & \\ 
g,e,l & 25,7,22 & & 35,4,15 & & 24,10,20 & & 21,12,21 & & 30,7,17 \\ 
\hline
RENN & 76.83 & \xmark & 76.99 & \xmark & 78.80 & \xmark & 78.20 & \xmark & 76.65 & \xmark \\ 
g,e,l & 32,3,19& & 35,4,15 & & 38,1,15 & & 35,2,17 & & 34,2,18 \\ 

\end{tabular}
\label{table:noNoise}
\end{table}

Examining the performance of the considered learning algorithms without noise handling (``orig'' in Tables \ref{table:weightingSchemes} and \ref{table:filtering}), we note that MLPs and random forests generally achieve the highest classification accuracy and may be the most tolerant to the inherent detrimental instances in each data set.
However, MLPs and Random Forests also appear to be the least robust to noise as they obtain the lowest average classification accuracy when more than 10\% of the instances are corrupted with noise.
With no artificial noise, MLPs and Random Forests achieve about 81\% accuracy.
With 20\% artificial noise, the average accuracy decreases to about 72\%.
On the other hand, C4.5, 5-NN, and RIPPER achieve an average accuracy of about 79\% with no artificial noise and an average accuracy of about 74\% with 20\% artificial noise.
With high degrees of noise, the built-in noise handling mechanisms of learning algorithms become more beneficial.

\subsection{Weighting Schemes}

As instance weighting is not as well explored as filtering, we now examine various weighting schemes to handle class noise.
Table \ref{table:weightingSchemes} compares RDIL-$\mathcal{L}$ with RDIL-Biased and PWEM.
The accuracies from the algorithms with no weighting are given to better measure the effectiveness of the methods.
RDIL-$\mathcal{L}$ significantly outperforms the other weighting schemes in most cases (represented by bold $\textit{p}$-values): 24 out of the 25 cases for PWEM, and 18 out of the 25 cases for RDIL-Biased.
In no case does a competing weighting scheme achieve significantly higher classification accuracy than RDIL-$\mathcal{L}$.
Recall that the nine learning algorithms were chosen to be diverse so as to represent more of the hypothesis space $\mathcal{H}$.
This suggests that a better estimation of $p(\hat{y}_i|x_i)$ produces better results for weighting and filtering.
This is shown empirically as RDIL-$\mathcal{L}$ and Filter-$\mathcal{L}$ have the most significant increase in accuracy for each learning algorithm (Table \ref{table:noNoise}).
However, there is an obvious trade-off since obtaining a more accurate estimate of $p(\hat{y}_i|x_i)$ is more computationally expensive.

\begin{table}
\centering
\caption{A comparison of the average accuracy from investigated instance weighting methods on the considered learning algorithms.
Bold values with a \cmark~represent cases where RDIL-$\mathcal{L}$ (R-$\mathcal{L}$) acheives significantly higher accuracy than RDIL-Biased (R-B) or PWEM (PW).}
\setlength{\tabcolsep}{0pt}
\centering
\begin{tabular}{l|ccccc||ccccc}
& \multicolumn{5}{c||}{C4.5}& \multicolumn{5}{c}{5-NN} \\
  & 0\% & 10\% & 20\% & 30\% & 40\% & 0\% & 10\% & 20\% & 30\% & 40\% \\
\hline          
R-$\mathcal{L}$ & 78.19 & 77.03 & 76.33 & 74.32 & 71.08 & 78.72 & 77.86 & 77.01 & 75.07 & 70.63 \\
\hline                     
R-B & 79.29 & 77.14 & \textbf{75.20}\cmark~ & \textbf{71.06}\cmark~ & \textbf{65.74}\cmark~ & \textbf{78.34}\cmark~& \textbf{77.41}\cmark~ & \textbf{75.39}\cmark~ & \textbf{71.59}\cmark~ & \textbf{64.92}\cmark \\
g,e,l & 25,4,25 & 32,2,19 & 38,0,16 & 44,0,10 & 43,1,10 & 34,7,13 & 41,3,9 & 43,1,10 & 46,1,7 & 44,1,9 \\
\hline
PW & \textbf{76.41}\cmark~ & \textbf{74.50}\cmark~ & \textbf{73.34}\cmark~ & \textbf{70.94}\cmark~ & \textbf{68.28}\cmark~ & \textbf{78.02}\cmark~ & \textbf{77.54}\cmark~ & \textbf{76.12}\cmark~ & \textbf{73.56}\cmark~ & \textbf{67.18}\cmark~ \\
g,e,l & 35,4,15 & 39,3,11 & 38,4,12 & 41,0,13 & 37,0,17 & 34,4,16 & 36,2,15 & 37,1,16 & 37,3,14 & 37,3,14 \\\hline
orig & 79.31 & 76.92 & 74.32 & 69.709 & 63.08 & 79.37 & 77.63 & 74.42 & 69.96 & 62.54\\
\multicolumn{11}{c}{}\\
&\multicolumn{5}{c}{MLP}&\multicolumn{5}{c}{Random Forest}\\                    
R-$\mathcal{L}$ & 82.26 & 80.7 & 78.40 & 75.17 & 69.30 & 80.82 & 79.72 & 78.06 & 75.89 & 70.78 \\\hline                     
R-B & \textbf{81.49}\cmark~ & \textbf{78.19}\cmark~ & \textbf{73.84}\cmark~ & \textbf{69.14}\cmark~ & \textbf{62.10}\cmark~ & 80.94 & \textbf{78.24}\cmark~ & \textbf{74.01}\cmark~ & \textbf{68.25}\cmark~ & \textbf{60.93}\cmark~ \\
g,e,l & 34,2,18 & 41,1,11 & 46,1,6 & 48,0,6 & 47,1,6 & 25,1,28 & 42,2,9 & 48,0,5 & 48,1,5 & 49,1,4 \\
\hline
PW & \textbf{82.79}\cmark~ & \textbf{79.67}\cmark~ & \textbf{76.42}\cmark~ & \textbf{71.9}\cmark~ & \textbf{65.95}\cmark~ & 81.51 &\textbf{78.69}\cmark~ & \textbf{76.37}\cmark~ & \textbf{72.54}\cmark~ & \textbf{65.87}\cmark~ \\
g,e,l & 34,4,16 & 37,2,14 & 38,1,14 & 42,0,12 & 42,0,12 & 33,4,17 & 34,4,15 & 40,4,9 & 47,1,6 & 48,1,5 \\\hline
orig & 81.67 & 77.46 & 72.25 & 67.17 & 60.46 & 81.18 & 77.75 & 72.72 & 66.87 & 59.63 \\
\multicolumn{11}{c}{}\\
&\multicolumn{5}{c}{RIPPER}\\ 
R-$\mathcal{L}$ & 77.86 & 76.54 & 75.54 & 73.46 & 69.63 \\ 
\cline{1-6}
R-B & 77.98 & 76.28 & \textbf{74.50}\cmark~ & \textbf{70.70}\cmark~ & \textbf{65.97}\cmark~ \\ 
g,e,l & 27,3,24 & 31,2,20 & 36,3,15 & 46,2,6 & 45,0,9 \\ 
\cline{1-6}
PW & \textbf{74.17}\cmark~ & \textbf{71.94}\cmark~ & \textbf{70.68}\cmark~ & \textbf{68.57}\cmark~ & \textbf{64.82}\cmark~ \\ 
g,e,l & 36,4,14 & 44,2,7 & 45,4,5 & 40,2,12 & 41,1,12 \\ \cline{1-6} 
orig & 78.35 & 76.32 & 73.45 & 69.87 & 65.10 \\
\end{tabular}
\label{table:weightingSchemes}
\end{table}

\subsection{Weighting VS Filtering}

We now compare weighting against filtering.
Weighting and filtering are {\em both} viable and significantly increase the classification accuracy when noise is added.
The difference between filtering and weighting techniques depends on the estimation of $p(y_i|x_i)$. 
Generally, estimating $p(y_i|x_i)$ with the set of learning algorithms $\mathcal{L}$ achieves greater classification accuracy than using a biased estimate.
Table \ref{table:filtering} compares RDIL-$\mathcal{L}$ with Filter-$\mathcal{L}$. 
With no noise, RDIL-$\mathcal{L}$ achieves significantly higher accuracy than Filter-$\mathcal{L}$ for the MLP.
Since each instance can affect the classification boundary for MLPs (as shown in Figure \ref{figure:example}), weighting the instances in the training set has a more significant impact in MLPs than the other learning algorithms which partition the input space. 
On the other hand, the Filter-$\mathcal{L}$ achieves a significantly higher accuracy than RDIL-$\mathcal{L}$ for the four other learning algorithms.
Note that MLP with RDIL-$\mathcal{L}$ achieves the highest overall average accuracy for noise levels 0\%-20\% (RDIL-$\mathcal{L}$ achieves the highest accuracy for 30\% and 40\% noise using Random Forest and C4.5 respectively).
Except for MLPs, the significance of the impact of RDIL-$\mathcal{L}$ increases as the noise level increases except for all of the examined learning algorithms.
RDIL-$\mathcal{L}$ significantly increases the classification accuracy for C4.5, 5-NN, and Random Forests when there are high amounts of noise.

Over all noise levels, RDIL-$\mathcal{L}$ compared with Filter-$\mathcal{L}$ achieves significantly higher classification accuracy in 6 of the 25 cases and the filter-$\mathcal{L}$ achieves significantly higher classification accuracy in 7 cases.
(In the other 12 cases, there is no significant difference).
The Filter-$\mathcal{L}$ has a more significant effect than RDIL-$\mathcal{L}$ for RIPPER at noise levels 0, 0.1 and 0.2 and RDIL-$\mathcal{L}$ never achieves significantly higher classification accuracy than the Filter-$\mathcal{L}$.
Therefore, instance weighting is not the best option for every learning algorithm.
However, with the Filter-$\mathcal{L}$ we chose the threshold that produced the highest classification accuracy on the test set, which is not always possible to do.
Instance weighting avoids the overhead of having to determine a threshold for filtering when using an ensemble filter.
Instance weighting is better for learning algorithms that consider each instance individually and each instance can affect the classification boundary (e.g. MLP).

\begin{table}
\centering
\caption{A comparison of RDIL-$\mathcal{L}$ (R-$\mathcal{L}$) with the $\mathcal{L}$-filter (F-$\mathcal{L}$) on the considered learning algorithms.
Bold values with a \cmark~represent cases where RDIL-$\mathcal{L}$ (R-$\mathcal{L}$) acheives significantly higher accuracy than the $\mathcal{L}$-filter.
The \xmark~represents cases where the $\mathcal{L}$-filter acheives significantly higher accuracy.}
\setlength{\tabcolsep}{0pt}
\begin{tabular}{l|ccccc||ccccc}
& \multicolumn{5}{c||}{C4.5}& \multicolumn{5}{c}{5-NN} \\  & 0\% & 10\% & 20\% & 30\% & 40\% & 0\% & 10\% & 20\% & 30\% & 40\% \\
\hline
R-$\mathcal{L}$ & 78.19 & 77.03 & 76.33 & 74.32 & 71.08 & 78.72 & 77.86 & 77.01 & 75.07 & 70.63 \\
\hline
F-$\mathcal{L}$ & 79.55~\xmark~ & 78.35~\xmark~ & 76.79 & 73.58 & \textbf{69.30}~\cmark~ & 79.40~\xmark~ & 78.35 & 76.6 & \textbf{74.35}~\cmark~ & \textbf{69.64}~\cmark \\
g,e,l & 19,2,33 & 21,1,31 & 30,1,23 & 33,0,21 & 36,0,18 & 19,4,31 & 28,2,23 & 26,4,24 & 31,0,23 & 31,0,23 \\\hline
orig & 79.31 & 76.92 & 74.32 & 69.709 & 63.08 & 79.37 & 77.63 & 74.42 & 69.96 & 62.54\\
\multicolumn{11}{c}{}\\
& \multicolumn{5}{c||}{MLP}& \multicolumn{5}{c}{Random Forest} \\
\hline
R-$\mathcal{L}$ & 82.26 & 80.7 & 78.4 & 75.17 & 69.30 & 80.82 & 79.72 & 78.06 & 75.89 & 70.78 \\
\hline
F-$\mathcal{L}$ & 81.80 & 80.66 & 78.24 & 74.85 & 69.46 & 81.66~\xmark~ & 79.91 & 78.06 & \textbf{75.29}~\cmark~ & \textbf{69.94}~\cmark \\
g,e,l & 37,2,15 & 30,3,20 & 31,1,21 & 33,0,21 & 29,1,24 & 18,4,32 & 31,0,22 & 26,1,26 & 33,0,21 & 34,3,17 \\\hline
orig & 81.67 & 77.46 & 72.25 & 67.17 & 60.46 & 81.18 & 77.75 & 72.72 & 66.87 & 59.63 \\
\multicolumn{11}{c}{}\\
& \multicolumn{5}{c||}{RIPPER} \\
\cline{1-6}
R-$\mathcal{L}$ & 77.86 & 76.54 & 75.54 & 73.46 & 69.63 \\
\cline{1-6}
F-$\mathcal{L}$ & 78.98~\xmark~ & 77.82~\xmark~ & 76.40~\xmark~ & 73.92 & 69.84 \\
g,e,l & 15,3,36 & 18,3,32 & 18,0,36 & 27,1,26 & 25,1,28 \\\cline{1-6}
orig & 78.35 & 76.32 & 73.45 & 69.87 & 65.10 \\
\end{tabular}
\label{table:filtering}
\end{table}

\section{Conclusions}
\label{section:conclusions}

In this paper we examined handling detrimental instances using the hypotheses from multiple learning algorithms. 
We introduced \textit{reduced detrimental instance learning} (RDIL) which weights each instance based on an approximation of $p(\hat{y}_i|x_i)$.
We examined RDIL on a set of 5 learning algorithms and 54 data sets.
We found that a better estimate of $p(\hat{y}_i|x_i)$ leads to better detrimentality handling in both instance weighting and filtering.
Weighting the instances avoids having to spend extra computational time and having to use training instances to select a threshold for filtering when using an ensemble filter.
Instance weighting has the greatest effect on learning algorithms where every instance can affect the classification boundary and the training instances are considered individually, such as multilayer perceptrons trained with backpropagation (MLPs).
On the other hand, instance filtering had a more significant impact on the C4.5, 5-NN, Random Forest, and RIPPER learning algorithms with no artificial noise.
However, instance weighting was shown to be preferable to filtering for the examined learning algorithms when there are high amounts of noise.
An analysis of when to use a particular noise handling technique is a direction for future work.
\bibliographystyle{IEEEtran}


\end{document}